# Experimental Comparison of Classification Uncertainty for Randomised and Bayesian Decision Tree Ensembles

V. Schetinin, D. Partridge, W.J. Krzanowski, R.M. Everson, J.E. Fieldsend, T.C. Bailey, and A. Hernandez

School of Engineering, Computer Science and Mathematics, University of Exeter, UK
{V.Schetinin, D.Partridge, W.J.Krzanowski, R.M.Everson,
J.E.Fieldsend, T.C.Bailey, A.Hernandez}@exeter.ac.uk

**Abstract.** In this paper we experimentally compare the classification uncertainty of the randomised Decision Tree (DT) ensemble technique and the Bayesian DT technique with a restarting strategy on a synthetic dataset as well as on some datasets commonly used in the machine learning community. For quantitative evaluation of classification uncertainty, we use an Uncertainty Envelope dealing with the class posterior distribution and a given confidence probability. Counting the classifier outcomes, this technique produces feasible evaluations of the classification uncertainty. Using this technique in our experiments, we found that the Bayesian DT technique is superior to the randomised DT ensemble technique.

## 1 Introduction

The uncertainty of classifiers used for safety-critical applications is of crucial importance. In general, uncertainty is a triple trade-off between the amount of data available for training, the classifier diversity and the classification accuracy [1 - 4]. The interpretability of classifiers can also produce useful information for experts responsible for making reliable classification, making Decisions Trees (DTs) an attractive scheme. The required diversity of classifiers can be achieved on the basis of two approaches: a DT ensemble technique [2] and an averaging technique based on Bayesian Markov Chain Monte Carlo (MCMC) methodology [3, 4]. Both DT techniques match the above requirements well and have revealed promising results when applied to some real-world problems [2 - 4].

By definition, DTs consist of splitting nodes and terminal nodes, which are also known as tree leaves. DTs are said to be binary if the splitting nodes ask a specific question and then divide the data points into two disjoint subsets called the left and the right branch. The terminal node assigns all data points falling in that node to the class whose points are prevalent. Within a Bayesian framework, the class posterior distribution having observed some data is calculated for each terminal node [3, 4].

The Bayesian generalization of tree models required to evaluate the posterior distribution of the trees has been given by Chipman *et al.* [3]. Denison *et al.* [4] have suggested MCMC techniques for evaluating the posterior distribution of decision trees. This technique performs a stochastic sampling of the posterior distribution.

In this paper we experimentally compare the classification uncertainty of the randomised DT ensemble technique and the Bayesian DT technique with a restarting strategy on a synthetic dataset and some domain problems from UCI Machine Learning Repository [5]. To provide quantitative evaluations of classification uncertainty, we use an Uncertainty Envelope dealing with the class posterior distribution and a given confidence probability [6]. Counting the classifier outcomes, this technique produces the feasible evaluations of the classification uncertainty.

Below in sections 2 and 3 we briefly describe the randomised and Bayesian DT techniques which are used in our experiments. Then in section 4 we briefly describe the Uncertainty Envelope technique used to quantitatively evaluate the uncertainty of the two classification techniques. The experimental results are presented in section 5, and section 6 concludes the paper.

## 2   The Randomised Decision Tree Ensemble Technique

Performance of a single DT can be improved by averaging the outputs of DTs involved in an ensemble [2]. The improvement is achieved if most of the DTs can correctly classify the data points misclassified by a single DT. Clearly, the required diversity of the classifier outcomes can be achieved if the DTs involved in an ensemble are independently induced from data. To achieve the required independence, Dietterich has suggested randomising the DT splits [2]. In this technique the best, in terms of information gain, 20 partitions for any node are calculated and one of these is randomly selected with uniform probability. The class posterior probabilities are calculated for all the DTs involved in an ensemble and then averaged.

A pruning factor, specified as the fewest number of data points falling in the terminal nodes, can affect the ensemble performance. However, within the randomised DT technique, this effect is insignificant when pruning does not exceed 10% of the number of the training examples [2]. More strongly the pruning factor affects the average size of the DTs, and consequently it has to be set reasonably.

The number of the randomised DTs in the ensemble is dependent on the classification problem and assigned by a user in an *ad hoc* manner. This technique permits the user to evaluate the diversity of the ensemble by comparing the performances of the ensemble and that of the best DT on a predefined validation data subset. The required diversity is achieved if the DT ensemble outperforms the best single DTs involved in the ensemble. Therefore this ensemble technique requires the use of *n*-fold cross-validation. In our experiments described in section 5 we used the above randomised DT ensemble technique. For all the domain problems the ensembles consist of 200 DTs. To keep the size of the DT acceptable, the pruning factor is set to be dependent on the number of the training examples. In particular, its value is set to 30 for problems with many training examples; otherwise it is 5. The performance of the randomised DT ensembles is evaluated on 5 folds for each problem.

## 3 The Bayesian Decision Tree Technique

In general, the predictive distribution we are interested in is written as the integral over parameters $\theta$ of the classification model

$$p(y \mid \mathbf{x}, \mathbf{D}) = \int p(y \mid \mathbf{x}, \theta, \mathbf{D}) p(\theta \mid \mathbf{D}) d\theta, \qquad (1)$$

where $y$ is the predicted class (1, …, $C$), $\mathbf{x} = (x_1, \ldots, x_m)$ is the $m$-dimensional input vector, and $\mathbf{D}$ is the data.

The integral (1) can be analytically calculated only in simple cases. Moreover, part of the integrand in (1), the posterior density of $\theta$ conditioned on the data $\mathbf{D}$, $p(\theta \mid \mathbf{D})$, cannot be evaluated except in very simple cases. However, if we can draw values $\theta^{(1)}, \ldots, \theta^{(N)}$ from the posterior distribution of $p(\theta \mid \mathbf{D})$, then we can write

$$p(y \mid \mathbf{x}, \mathbf{D}) \approx \sum_{i=1}^{N} p(y \mid \mathbf{x}, \theta^{(i)}, \mathbf{D}) p(\theta^{(i)} \mid \mathbf{D}) = \frac{1}{N} \sum_{i=1}^{N} p(y \mid \mathbf{x}, \theta^{(i)}, \mathbf{D}) \cdot \qquad (2)$$

This is the basis of the MCMC technique for approximating integrals [3, 4]. To perform the approximation, we need to generate random samples from $p(\theta \mid \mathbf{D})$ by running a Markov Chain until it has converged to the stationary distribution. After this we can draw samples from the Markov Chain and calculate the predictive posterior density (2). For integration over models in which the dimension of $\theta$ varies, MCMC methods permit Reversible Jumps as described in [4].

Because DTs are hierarchical structures, changes at the nodes located at the upper levels (close to their roots) can cause drastic changes to the location of data points at the lower levels. For this reason there is a very small probability of changing and then accepting a DT located near a root node. Therefore RJMCMC algorithms tend to explore the DTs in which only the splitting nodes located far from the root node are changed. These nodes typically contain small numbers of data points. Consequently, the value of the likelihood is not changed much, and such moves are always accepted. As a result, RJMCMC algorithms cannot explore a full posterior distribution.

The space which is explored can be extended by using a *restarting strategy* as Chipman *et al.* have suggested in [3]. The idea behind the restarting strategy is based on multiple runs of the RJMCMC algorithm with short intervals of burn-in and post burn-in. For each run, the algorithm creates an initial DT with the random parameters and then starts exploring the tree model space. Running short intervals prevents the DTs from getting stuck at a particular DT structure. More important, however, is that the multiple runs allow the exploring of the DT model space starting with very different DTs. So, averaging the DTs over all such run can improve the performance of the RJMCMC algorithm. The disadvantage, of course, is that the multiple short chains with short burn-in runs, will seldom reach a stationary distribution. The restarting strategy, as we see, does not limit the DT sizes explicitly, as would be done by a restricting strategy [4]. For this reason the restarting strategy seems to be the more practical. In section 4 we use this strategy in our comparative experiments. The quantitative comparison of the classification uncertainty is done within the Uncertainty Envelope technique described next.

## 4 The Uncertainty Envelope Technique

Let us consider a simple example of a classifier system consisting of $N = 1000$ classifiers in which 2 classifiers give a conflicting classification. Then for a given datum **x** the posterior probability $P_i = 1 - 2/1000 = 0.998$. In this case we can conclude that the multiple classifier system was trained well and/or the datum **x** lies far from the class boundaries. For this datum, and for each new data point appearing in some neighbourhood of the datum **x**, the classification uncertainty as the probability of misclassification is expected to be $1 - P_i = 0.002$. For other data points, the values of $P$ differ and range between $P_{min}$ and 1. It is easy to see that $P_{min} = 1/C$.

When the value of $P_i$ is close to $P_{min}$, the classification uncertainty is highest and a datum **x** can be misclassified with a probability $1 - P_i = 1 - 1/C$. So we can assume some value of probability $P_0$ for which the classifier outcome is expected to be confident, that is the probability with which a given datum **x** could be misclassified is small enough to be acceptable. Given such a value of $P_0$, we can now specify the confidence or, *vice versa*, the uncertainty of classifier outcomes in statistical terms. The classification outcome is said to be *confidently correct*, when the probability of misclassification is acceptably small and $P_i \geq P_0$.

Additionally to the confidently correct output, we can specify a *confidently incorrect* output referring to a case when almost all the classifiers assign a datum **x** to a wrong class $j$, i.e., $P_j \geq P_0$. By definition this evaluation tells us that most of the classifiers fail in the same manner to classify a datum **x**. This can happen for different reasons, for example, the datum **x** could be mislabelled or corrupted, or the classifiers within a predefined scheme cannot properly distinguish such data points.

The remaining cases, for which $P_i < P_0$, are regarded as *uncertain classifications*. In such cases the classifier outcomes cannot be accepted with a given confidence probability $P_0$. The multiple classifier system labels such outcomes as uncertain.

The above three characteristics, confidently correct, confidently incorrect, and uncertain outcomes, seem to provide a good way of evaluating different types of multiple classifier systems on the same data. Comparing the values of these characteristics, we can quantitatively evaluate the classification uncertainty of these systems. Depending on the costs of types of misclassifications in real applications, we have to specify the value of the confidence probability $P_0$, say $P_0 = 0.99$.

## 5 Experiments and Results

First we conduct experiments on synthetic dataset and then on 7 domain problems taken from the UCI Repository. A two dimensional synthetic dataset was generated as a mixture of five Gaussians. The data points drawn from the first three Gaussians belong to class 1 and the data points from the remaining two Gaussians to class 2.

The mixing weights $\rho_{ij}$ and kernel centres $\mu_{ij}$ of these Gaussians for class 1 are $\rho_{11} = 0.16$, $\mu_{11} = (1.0, 1.0)$, $\rho_{12} = 0.17$, $\mu_{12} = (0.7, 0.3)$, $\rho_{13} = 0.17$, $\mu_{13} = (0.3, 0.3)$ and for class 2 they are $\rho_{21} = 0.25$, $\mu_{21} = (-0.3, 0.7)$, $\rho_{22} = 0.25$, $\mu_{22} = (0.4, 0.7)$. The kernels all have isotropic covariance: $\Sigma_i = 0.03\mathbf{I}$. This mixture is an extended version of the

Ripley data [7]. Because the classes overlap, the Bayes error on these data is 9.3%. 250 data points drawn from the above mixture form the training dataset. Another 1000 data points drawn from the mixture form the testing data.

Both the randomised DT ensemble and the Bayesian techniques were run on the above synthetic data. The pruning factor was set equal to 5. On the synthetic data, the ensemble output quickly converges and stabilizes after averaging 100 DTs. The mean size of DTs and the standard deviation over all 5 folds were 32.9 and 3.3, respectively. The averaged classification performance was 87.12%. Within the Uncertainty Envelope, the rates of confidently correct, uncertain, and confidently incorrect outcomes were 78.9%, 9.8%, and 11.3%, respectively. The widths of $2\sigma$ intervals for these outcomes were 34.9%, 43.7%, and 8.9%, respectively. We can see that the values of the intervals calculated for the confidently correct and the uncertain outcomes are very large. This happens because the randomised DT ensemble technique produces mostly uncertain outcomes on some of the 5 folds.

Using the restarting strategy, the Bayesian DTs were run 50 times; each time 2000 samples were taken for burn-in and 2000 for post burn-in. The probabilities of birth, death, change variable, and change rule were 0.1, 0.1, 0.1, and 0.7, respectively. Uniform priors on the number of inputs and nodes were used [4]. The resultant average classification performance was 87.20%, and the mean size and the standard deviation of the DTs were 12.4 and 2.5, respectively. The rates of confidently correct, uncertain and confidently incorrect outcomes were 63.30%, 34.40% and 2.30%, respectively. So we can see, first, that on the synthetic data the Bayesian DTs are much shorter than those of the randomised ensemble. Second, the randomised DT ensemble technique cannot provide reliable classifier outcomes. Of course, the 5 fold data partition used in the randomised DT technique makes an additional contribution to the classification uncertainty. However, practically this effect disappears for domain problems including more than 300 data points.

Table 1 lists the characteristics of the 7 domain problems used in our experiments; here *C*, *m*, *train*, and *test* are the numbers of classes, input variables, training and testing examples, respectively. This table also provides the performances of the Bayesian DT technique on these data.

**Table 1.** Performances of the Bayesian DTs with restarting strategy

| Data | Data characteristics | | | | DT size | Perform, % | Uncertainty Envelope, % | | |
|---|---|---|---|---|---|---|---|---|---|
| | C | m | train | test | | | Correct | Uncertain | Incorrect |
| *Ionosphere* | 2 | 33 | 200 | 151 | **11.99±2.2** | **95.35** | 11.92 | 88.08 | **0.00** |
| *Winconsin* | 2 | 9 | 455 | 228 | **11.81±2.4** | **99.12** | 82.89 | 17.11 | **0.00** |
| *Image* | 7 | 19 | 210 | 2100 | **15.71±2.3** | 94.29 | 22.38 | 77.62 | **0.00** |
| *Votes* | 2 | 16 | 391 | 44 | **10.25±2.5** | 95.45 | 56.82 | 43.18 | **0.00** |
| *Sonar* | 2 | 60 | 138 | 70 | **9.94±1.8** | **81.43** | 0.00 | 100.00 | **0.00** |
| *Vehicle* | 4 | 18 | 564 | 282 | **47.78±4.6** | 69.86 | 3.90 | 96.10 | **0.00** |
| *Pima* | 2 | 8 | 512 | 256 | **11.99±22** | 79.69 | 34.77 | 60.55 | **4.69** |

The performances of the randomised technique are shown in Table 2. This table shows also the classification performance of the best single DTs selected on the validation subsets averaged over all the 5 folds. From Table 2, we can see that the randomised DT ensemble technique always outperforms the best single DTs.

Comparing the randomised and Bayesian DT ensembles, we can conclude that on all the datasets the Bayesian DTs are shorter by 2 to 3 times those of the randomised ensembles. Both ensembles have the same performance on the Image, Votes, Vehicle, and Pima datasets. However, on the remaining datasets, the Bayesian technique slightly outperforms the randomised ensemble technique.

It is interesting to note the Bayesian ensemble method always makes a smaller proportion of confidently correct classifications (although, variances from the randomised ensemble are very high). Likewise, the proportion of confidently incorrect classifications is always higher for the randomised ensemble. Indeed, on the synthetic data, the randomised ensemble classifiers on average make more confidently incorrect classifications (11.3%) than the Bayes error rate (9.3%), whereas the Bayesian ensemble makes only 2.3% confidently incorrect classifications.

In fact, as Table 1 shows, the Bayesian DTs seldom make confident, but incorrect classifications, though they make more uncertain classifications. Although it may be unrealistic to expect the confidently incorrect rate to approach the Bayes error rate with small datasets, these results suggest that the randomised ensemble tends to produce over-confident ensembles, while the Bayesian ensembles make few confident but incorrect classifications.

On the other hand, as exemplified by the Ionosphere and Sonar datasets, the Bayesian ensemble may yield accurate classifications, but the majority of them may be uncertain. The Sonar and Ionosphere data have 60 and 33 features respectively and relatively few data points, so it is unsurprising that the sparsity of data points in these high-dimensional datasets leads to uncertain classifications.

Table 2. Performances of the randomised DT ensembles

| Data | Single DT perform, % | DT size | Perform, % | Uncertainty Envelope, % | | |
|---|---|---|---|---|---|---|
| | | | | Correct | Uncertain | Incorrect |
| *Ionosphere* | 88.8±8.0 | 21.2±1.3 | 94.4±0.7 | **76.5**±35.8 | 7.0±44.4 | 16.5±18.4 |
| *Winconsin* | 96.1±1.7 | 32.7±1.5 | 97.7±1.2 | 96.7±7.90 | 1.4±9.2 | 1.9±1.8 |
| *Image* | 87.4±4.4 | 27.9±1.3 | 94.2±0.9 | 86.1±33.0 | 6.5±37.9 | 7.4±7.9 |
| *Votes* | 93.9±3.1 | 27.1±3.6 | 95.2±1.4 | 94.3±5.80 | 1.1±7.2 | 4.5±2.1 |
| *Sonar* | 70.7±7.8 | 17.8±0.8 | 78.3±5.5 | **54.9**±40.6 | 9.6±60.5 | 35.6±31.8 |
| *Vehicle* | 69.0±4.5 | 115.8±3.2 | 71.9±2.2 | **63.8**±31.0 | 8.8±50.2 | 27.4±20.1 |
| *Pima* | 77.3±1.2 | 33.6±4.0 | 80.2±2.4 | 66.7±47.0 | 14.6±65.3 | 18.7±19.6 |

## 6 Conclusion

We have experimentally compared the classification uncertainty of the randomised DT ensemble technique with the ensembles sampled from the Bayesian posterior using RJMCMC with a restarting strategy. The ensemble techniques both outperform the best single DT, having similar average classification rates. Far fewer confidently incorrect classifications are made by the Bayesian ensemble. This is clearly a very desirable property for classifiers in safety-critical applications in which confidently made, but incorrect, classifications may be fatal.

## Acknowledgments

This research was supported by the EPSRC, grant GR/R24357/01.